\documentclass[conference,compsoc]{IEEEtran}
\usepackage{cite}
\usepackage{amsmath,amssymb,amsfonts}
\usepackage{algorithmic}
\usepackage{graphicx}
\usepackage{textcomp}
\usepackage{xcolor}
\usepackage{booktabs}
\usepackage{xurl}
\usepackage[most]{tcolorbox}

\graphicspath{{figures/}}
\newtcolorbox{finding}{colback=white,colframe=gray!55,arc=1pt,boxrule=0.3pt,left=8pt,right=8pt,top=6pt,bottom=6pt,before skip=8pt,after skip=8pt}
\def\BibTeX{{\rm B\kern-.05em{\sc i\kern-.025em b}\kern-.08emT\kern-.1667em\lower.7ex\hbox{E}\kern-.125emX}}
\begin{document}

\title{FOE-Dreamer: Deployment-Efficient Learning of Cyber Defense Policies in Operational Networks
}

\author{\IEEEauthorblockN{Yinuo Du}
\IEEEauthorblockA{\textit{Department of Computer Science} \\
\textit{University of Texas at El Paso}\\
El Paso, TX, USA \\
aduyino6@gmail.com}
\and
\IEEEauthorblockN{Christopher Kiekintveld}
\IEEEauthorblockA{\textit{Department of Computer Science} \\
\textit{University of Texas at El Paso}\\
El Paso, TX, USA \\
cdkiekintveld@utep.edu}
}

\maketitle

\begin{abstract}
Autonomous cyber defense is widely studied with deep reinforcement learning, but most work trains policies inside simulators because learning requires many trial-and-error interactions. Simulator-based training carries a fidelity gap: trained policies face authentication delays, partial action failures, service dependencies, and user traffic patterns that the simulator did not represent, and degrade on transfer. Training directly in the network being defended removes the gap but makes every interaction slow and costly, so sample efficiency becomes the binding constraint. Model-based reinforcement learning has reduced sample requirements on games and physical robots by orders of magnitude by planning inside a learned world model, but it has not been applied to defenders trained in operational networks. We present FOE-Dreamer (Factored Opponent-Enhanced Dreamer), a model-based reinforcement learning framework that trains a defender online and directly in a network on cloud infrastructure, without a simulator stage. FOE-Dreamer separates three sources of variation in the latent state: a controllable component driven by defender actions, an exogenous component driven by benign user activity, and an opponent embedding inferred from defender-observable signals alone. Separating the components reduces the variance in policy gradients during planning in imagination. Against two scripted attacker profiles with user activity emulated by the GHOSTS framework on an 8-host OpenStack testbed, FOE-Dreamer trains within a 3-day budget on a single GPU and achieves about 2$\times$ better episode loss than Rainbow and IQN baselines under matched compute. Both the factoring and the opponent model contribute to the gain, and the learned latents are consistent with the intended separation.
\end{abstract}

\begin{IEEEkeywords}
autonomous cyber defense, model-based reinforcement learning, opponent modeling, world models, cyber deception
\end{IEEEkeywords}

\section{Introduction}

Operators defending enterprise networks face attacks that move at machine timescale. Adversaries chain reconnaissance, exploitation, lateral movement, and exfiltration in minutes. Human analysts cannot keep pace, so the community has turned to reinforcement learning (RL) to build defenders that act on their own \cite{vyas2025acnd}. Most RL defenders learn inside a simulator and then transfer the policy to the network they protect. Simulators abstract away authentication delays, partial action failures, service dependencies, and the rhythm of legitimate user traffic \cite{kiely2025cage4}. The fidelity gap between simulator and operational network is the recurring obstacle to deployable autonomous defense \cite{mcfadden2026sok}.

The term ``operational network'' in this paper means an emulated multi-host enterprise environment that runs real software on a real cloud substrate, with sub-10\,s polling and behaviorally correlated user activity. It does not mean a production network with real attackers and real users.

Existing autonomous-defense RL falls into three groups. The first trains in a high-fidelity simulator and transfers, including the CybORG and CAGE Challenge lineage \cite{standen2021cyborg,kiely2025cage4}, Cyberwheel \cite{oesch2024cyberwheel}, and Yawning Titan \cite{acuto2023yawningtitan}. The second trains in a low-fidelity simulator and accepts the gap, including entity-based gyms such as CyberBattleSim \cite{microsoft2021cyberbattlesim} and NetSecGame \cite{rigaki2024outofcage}. The third avoids RL and plans over hand-specified models, including Mirage \cite{kouremetis2024mirage} and classical playbook-driven systems. None of these survive contact with an operational network that runs real services, real CVEs, and concurrent user activity. Transfer experiments report success rates as low as 66 percent when policies move from simulation to emulation \cite{standen2021cyborg}. The Vyas et al.\ autonomous cyber-defense survey names the realism of agent training environments and sim-to-real transfer as the field's open problems \cite{vyas2025acnd}.

The tradeoff between simulator fidelity and per-interaction cost can be read as a ladder. Abstract simulators are inexpensive per interaction but unrealistic. Production deployment is realistic but each exploratory action carries operational cost. Higher-fidelity emulation sits in the middle: real software, real CVE chains, and behaviorally correlated user activity, at higher per-interaction cost than abstract simulation but still tractable. Bauer et al.\ identify this middle rung, with in-network training, as the path forward for defender systems at machine timescale \cite{bauer2026cyber}. FOE-Dreamer trains a defender directly in such an emulated operational network, with no separate simulator stage. Policy optimization still uses simulations in the form of imagined latent rollouts from the learned world model.

Two architectural choices support training under the resulting sample budget. First, a factored world model separates defender-driven and exogenous latent dynamics to reduce confounding from background user activity. Second, an opponent model infers attacker type from defender-observable signals and conditions policy planning on the inferred type. Most policy updates happen on imagined rollouts in the learned latent space, so real-network interaction is reserved for data collection. Section~\ref{sec:scenario} traces a concrete intrusion that exposes the three challenges these choices target.

On an 8-host enterprise OpenStack testbed with two scripted attacker profiles, Avoidant and Enticed, and user activity generated by the GHOSTS user-emulation framework \cite{updyke2018ghosts}, FOE-Dreamer trains for three days on a single GPU and reduces episode loss against both profiles by roughly $2\times$ relative to Rainbow and IQN. Ablations show that both the factoring and the opponent model contribute to the gain.

We make three contributions. The first is a factored world model that disentangles defender-driven, user-driven, and attacker-driven sources of variation in the latent state, reducing the spurious credit assignment that monolithic-state RL incurs in operational networks. The second is an opponent model that infers attacker type from local defender observations and conditions imagined rollouts on the inferred type, supporting earlier and more accurate containment decisions against probe-and-withdraw adversaries. The third is the supporting training environment: among prior autonomous cyber defense environments, none combines the OpenStack substrate, real-CVE exploits, GHOSTS-class user emulation, and fine-grained wall-clock RL polling in a single artifact; FOE-Dreamer's testbed does.\footnote{Code, configurations, and reproduction recipes are available at \url{https://anonymous.4open.science/r/FOE-DREAMER-5117}.}

\section{Background and Related Work}

\subsection{Cyber Deception}

Cyber deception places artifacts in a network --- honeypots, honeytokens, decoy services --- to mislead attackers, slow their progress, and reveal their presence \cite{prebot2023learning}. Vetterl and Clayton show that low- and medium-interaction honeypots can be fingerprinted through differences in protocol implementations \cite{vetterl2018bitterharvest}. The effect of a decoy therefore depends on whether an attacker avoids or continues interacting with it. Recent work studies adaptive deception, where placement and configuration are chosen by a learned policy responding to observed attacker behavior \cite{sayed2023honeypot,du2022adaptive}. FOE-Dreamer's defender includes deception primitives --- a Cowrie SSH honeypot and a fake set-user-ID (SUID) binary --- among its actions, and learns when and where to deploy them based on observed network state.

\subsection{Autonomous Cyber Defense and the Simulator Gap}

Autonomous cyber network defense (ACND) trains agents to detect, contain, and recover from intrusions without a human in the loop. The survey by Vyas et al.\ provides a canonical taxonomy and inventories more than twenty ACND environments \cite{vyas2025acnd}. The dominant simulators are CybORG and the CAGE Challenge series \cite{standen2021cyborg,kiely2025cage4}, Yawning Titan \cite{andrew2022yawningtitan}, CyberBattleSim \cite{microsoft2021cyberbattlesim}, and NetSecGame \cite{rigaki2024outofcage}. Each abstracts a different slice of the network: CybORG models discrete kill-chain steps, CyberBattleSim models lateral movement on an attack graph, NetSecGame models exfiltration with stochastic detection. None capture authentication latencies, real exploit chains, and human-behavior-correlated user traffic at the same time.

The simulator-fidelity gap is the recurring failure mode of this lineage \cite{vyas2025acnd,mcfadden2026sok}. Transfer experiments support the concern: the original CybORG report observed only a 66\% success rate when policies trained in simulation were evaluated on the matching emulator \cite{standen2021cyborg}. The Vyas survey lists realistic agent training environments and sim-to-real transfer as the field's two largest open problems \cite{vyas2025acnd}. FOE-Dreamer collects experience and evaluates policies in the emulated network; its learned world model supplies imagined rollouts for policy optimization.

\subsection{Adversary Modeling}

Effective defense depends on inferring attacker behavior from defender-observable signals. Classical opponent modeling formalizes this inference inside interactive partially observable Markov decision process (POMDP) models or game-theoretic security games, but these methods typically require known opponent type spaces or access to opponent trajectories \cite{albrecht2018autonomous,gmytrasiewicz2005framework,nguyen2013analyzing}. Representation-learning approaches sidestep both requirements by learning latent embeddings of opponent behavior directly from data: He et al.\ \cite{he2016opponent} introduced opponent modeling with deep RL; Grover et al.\ \cite{grover2018learning} proposed encoder-decoder architectures for policy representations; Rabinowitz et al.\ \cite{rabinowitz2018machine} introduced Theory of Mind networks for meta-learning over modeled agents. Most early methods assume access to opponent trajectories during execution, which is rarely available in cybersecurity, where defenders see only local observations. Papoudakis et al.\ \cite{papoudakis2021agent} relaxed this assumption by learning the relationship between the controlled agent's trajectory and the modeled agents' trajectories. Recent work has explored richer prediction targets and transformer-based architectures \cite{jing2024towards}. FOE-Dreamer adopts this encoder-decoder pattern \cite{papoudakis2021agent}: its recurrent encoder reads recent defender observations and actions, and the learned embedding conditions policy optimization and is recomputed during imagined rollouts.

\subsection{Model-Based RL and World Models}

Model-free RL needs millions of environment interactions to converge, which is infeasible when each interaction touches a running network \cite{salvato2021reality}. Model-based RL learns a world model of environment dynamics and improves the policy through imagined rollouts in latent space, reducing the number of real interactions by orders of magnitude \cite{ha2018worldmodels,hafner2020dreamer,hafner2021dreamerv2,hafner2023dreamerv3}. The Dreamer family has matched or beaten model-free baselines on Atari, robotic locomotion, and Minecraft using a fraction of the samples \cite{hafner2023dreamerv3}.

Prior model-based RL in security is sparse, and what exists is on the attacker side. Rigaki and Garcia's MEME uses a learned surrogate of an antivirus classifier to drive sample-efficient malware evasion \cite{rigaki2023meme}. To our knowledge, no prior work applies a Dreamer-style world model on the defender side or trains a world-model agent in an operational network. Bauer et al.\ identify this gap as a foundational obstacle for defender systems at machine timescale \cite{bauer2026cyber}.

\subsection{Reinforcement Learning for Cyber Defense}

Reinforcement learning is widely used in autonomous cyber defense \cite{li2021cygil}. Proximal policy optimization is a frequent baseline, applied for example to adaptive cyber deception games \cite{du2022adaptive}. Hierarchical RL has been used for advanced-persistent-threat defense, and curiosity-driven exploration for discovering novel defensive strategies \cite{foley2022autonomous,singh2025hierarchical}. RL defenders are vulnerable to overfitting on specific opponent behaviors \cite{cobbe2019quantifying}, which motivates opponent-aware training. Sample efficiency is the persistent constraint: model-free algorithms require millions of interactions that are infeasible in operational networks where each interaction carries operational cost \cite{nguyen2022deep}. Prior model-based-RL work in cyber defense has relied on simulators such as CybORG \cite{standen2021cyborg}, where the sim-to-real gap remains a concern.

\subsection{Operational Training Environments}

A small number of recent systems train or evaluate cyber agents on substrates closer to operational networks. MHBench and Incalmo build OpenStack-backed multi-host benchmarks for LLM red teams, with real CVEs but no user emulation and no RL interface \cite{singer2025incalmo}. The Cyber Security Learning Environment (CSLE) provides a Docker-based emulation with CVE-tagged containers and an RL polling loop, but not OpenStack and not GHOSTS-class user behavior \cite{hammar2023csle}. Cyberwheel offers an emulator on FIREWHEEL/QEMU with CVE-tagged hosts, but does not integrate GHOSTS users \cite{oesch2024cyberwheel}. Kyoushi provides state-machine user emulation closest in spirit to GHOSTS but is a testbed builder, not an RL gym \cite{landauer2020kyoushi}. GHOSTS itself produces behaviorally-correlated user traffic but ships no defender RL stack \cite{updyke2018ghosts}.

CSLE learns strategies in simulation using statistics measured in emulation, then evaluates them in emulation \cite{hammar2023csle}. FOE-Dreamer uses imagined latent rollouts and refits its world model throughout training.

Among prior autonomous cyber defense environments, none combines the OpenStack substrate, real-CVE exploits, GHOSTS-class user emulation, and fine-grained wall-clock RL polling. The closest analogues are MHBench/Incalmo on the substrate-and-CVE axes, CSLE on the CVE-and-polling axes, and Kyoushi on the user-emulation axis. FOE-Dreamer's testbed integrates all four, as detailed in Section~\ref{sec:testbed}.

\section{Motivating Scenario}\label{sec:scenario}

The defender can isolate, reboot, monitor, deploy deception, or wait (Isolate, Disrupt, Monitor, Deceive, Null; Section~\ref{sec:defender}). Avoidant withdraws after suspicious probe results; Enticed may continue for valuable opportunities (Section~\ref{sec:threat-model}).

Consider a mid-size enterprise whose perimeter looks like the network in our testbed: two public-facing WordPress hosts (PublicWeb1, PublicWeb2), a private subnet with an NTP server and a database server, and an internal subnet with one web server and three user workstations. The shape echoes the Equifax breach, where an unpatched Apache Struts host (CVE-2017-5638) opened the door to 48 internal database hosts and the exfiltration of records on 147 million people. We trace one episode in our environment to show what the defender sees, what the defender does not see, and why that gap is hard.

\textbf{Initial access.} An external scanner reaches PublicWeb1 and exploits the WordPress Backup Migration vulnerability (CVSS 9.8) \cite{cve20236553}. The exploit yields a PHP-level shell. PublicWeb1 emits a single intrusion-detection alert that looks like every other malformed request the server saw that morning. The defender sees Alerts Count rise by one on PublicWeb1.

\textbf{Kill chain through the network.} From the foothold, the intruder enumerates services on the private subnet (port 22 on the NTP and database servers, 5432 on the database), tries the NTP server's weak SSH credentials, a vulnerability class listed in Table~\ref{tab:vuln}, and opens a user-level shell on a successful guess. The intruder then lists \texttt{/etc/cron.d}, finds a world-writable script that runs as root, rewrites it in place, and waits for the next cron tick to gain root. With root on the NTP server, the intruder harvests credentials from \texttt{/root/.pgpass}, pivots to the database server, and dumps tables for outbound transfer. Throughout, concurrent user activity overlaps the intruder's traces: an administrator's scripted health check from PC2 produces a port scan indistinguishable from reconnaissance; a backup job from PC1 opens an SSH session to the NTP server overlapping the attacker's session in the host log; the internal web server issues legitimate database queries during exfiltration. Compromise and escalation are not directly observable in the defender's feature vector. Alert volume rises on the affected hosts, but each rise has a plausible benign explanation.

Throughout the episode, GHOSTS-generated user activity continues on every host: HTTP browsing on the public servers, SSH admin sessions on the NTP and database hosts, periodic database queries from the internal web server, and file transfers between PCs. The defender polls every 5\,s and selects one of these actions. Isolating PublicWeb1 early would have stopped the attack at the cost of dropping active user sessions. Doing nothing leaves the database open. Deploying a Cowrie SSH honeypot on the NTP server would have caught the attacker if the attacker's profile is Enticed, and would have signaled ``decoy'' and triggered withdrawal if the profile is Avoidant.

This episode exposes three challenges that recur throughout the rest of the paper.

\textbf{C1. Entangled observations.} Every signal the defender sees --- Alerts Count, Honeypot Status, Host Availability, Trap Indicators, exploit and decoy timestamps --- is the joint product of attacker behavior and concurrent user behavior. A naive defender attributes legitimate diagnostic scans to the intruder and isolates the wrong host, or attributes the intruder's reconnaissance to a noisy admin and waits too long. Sample-efficient learning requires separating user-driven dynamics from attacker-driven dynamics in the world model.

\textbf{C2. Profile-dependent response.} The same observation sequence calls for different responses against different attackers. An Avoidant intruder withdraws on the first deception indicator, so cheap honeypots are highly effective and aggressive isolation is wasteful. An Enticed intruder ignores ambiguous cues, so deception buys little and containment must be decisive. The defender must infer the profile from limited, noisy observations rather than assume it.

\textbf{C3. Cost of training in the operational network.} Each interaction touches running VMs. Isolation drops user traffic. Reboots clear attacker persistence and disrupt active sessions. A model-free defender that needs millions of episodes to converge is not deployable in this regime. Sample efficiency, not asymptotic performance, is the central constraint.

These challenges motivate the three components of FOE-Dreamer described in Section~\ref{sec:approach}: a factored world model for the entanglement in C1, an opponent model for the profile uncertainty in C2, and model-based RL with imagined rollouts for the sample budget in C3.

\section{Threat Model and Testbed}\label{sec:testbed}

\subsection{Threat Model}\label{sec:threat-model}

FOE-Dreamer defends an enterprise network against a post-initial-access intruder. The attacker is assumed to have already breached the perimeter --- through phishing, a public-facing exploit, or a compromised credential --- and is now operating from a foothold on a publicly reachable host. The defender's job is to contain expansion before exfiltration, not to prevent the initial breach.

The attacker's action space is a six-element set drawn from the post-delivery phases of the kill chain: Discover (enumerate hosts, services, vulnerabilities), Compromise (exploit a known vulnerability), Escalate (elevate from user to root), Exfiltrate (extract data from a compromised host), Detect Deception (probe for honeypot or decoy artifacts), and Null (no action this step). At each timestep the attacker selects one action consistent with its current state. We model two fixed-class profiles that bracket the spectrum of attacker sophistication observed in field studies. Both profiles use three deterministic binary probes that evaluate deception indicators using only attacker-observable information: (i) a banner-fingerprint probe that compares a service handshake against expected production signatures, operationalising the systematic honeypot fingerprinting techniques demonstrated by Vetterl and Clayton \cite{vetterl2018bitterharvest}; (ii) a credential-validation probe that performs a single authentication attempt against a canonical service when apparent credentials are discovered; and (iii) a post-exploit verification probe that runs a command on a newly obtained shell and checks for expected production markers. Each probe returns \textsc{Suspicious} or \textsc{Not\_Suspicious}; the two profiles differ in the decision rule applied to those outcomes.

\textbf{Avoidant attacker.} The Avoidant attacker applies a conservative decision rule: any \textsc{Suspicious} outcome from any probe triggers immediate withdrawal from the affected host. This models sophisticated adversaries who prioritize operational security and long-term persistence, consistent with the APT-style behavior documented in field studies of cyber deception \cite{fergusonwalter2019tularosa,fergusonwalter2021decoy}. Banner inconsistencies cause host abandonment; a failed credential validation causes the host to be classified as compromised and retreated from; missing post-exploit markers trigger immediate session termination.

\textbf{Enticed attacker.} The Enticed attacker applies a permissive decision rule, tolerating \textsc{Suspicious} outcomes when high-value opportunities are present. This models opportunistic adversaries including automated malware, insider threats, and financially motivated actors who prioritize rapid value extraction over stealth, consistent with the ``smash-and-grab'' behavior of commodity attacks \cite{antonakakis2017mirai,barron2017picky}. Banner anomalies are disregarded and exploitation proceeds; a failed credential causes only that artifact to be skipped while exploration continues; a failed post-exploit verification shifts priority to other targets rather than triggering retreat.

Several threats are out of scope. We do not consider adaptive learning adversaries that retrain against the deployed defender; the two profiles are scripted and fixed-class. We do not consider attacks on the machine-learning components of the defender itself: no model extraction, no policy poisoning, no adversarial-example attacks on the world model. We do not consider zero-day exploits beyond the four CVE-class vulnerabilities listed in Table~\ref{tab:vuln}. We do not consider insiders with legitimate credentials whose actions are indistinguishable from authorized use. Defenses orthogonal to ours --- perimeter firewalls, network IDS, endpoint EDR, multi-factor authentication --- are neither required nor excluded. The threat model is consistent with prior work on autonomous cyber network defense \cite{rigaki2024hackphyr,rigaki2024outofcage}.

\subsection{Defender Capability}\label{sec:defender}

The defender observes a per-host feature vector at a 5-second polling interval. Each host contributes Alerts Count (number of IDS or host-log alerts since the previous poll), Honeypot Status (whether a honeypot is active on the host), Host Availability (up or down), Trap Indicators (whether any deployed trap has fired), and timestamped events for exploit attempts and decoy interactions. The defender does not directly observe whether a host is compromised, whether privilege escalation has occurred, what action the attacker took, or what the attacker intends next. All inferences about attacker state must be drawn from the observable vector.

The defender's action space is five-element. Isolate quarantines a host by changing its OpenStack security-group membership; the host becomes unreachable from other subnets and from users. Disrupt reboots a host over SSH; this clears in-memory persistence but terminates active user sessions. Monitor raises log verbosity on a target host, increasing future Alerts Count fidelity at no operational cost to users. Deceive has two variants: activate a Cowrie SSH honeypot bound to weak credentials on a target host, or plant a fake SUID binary intended to attract privilege-escalation attempts. Null lets the timestep pass without intervention. An episode runs for 100 timesteps, roughly eight minutes of wall-clock time. Reward penalizes successful attacker progression (compromise, escalation, exfiltration) and penalizes user-visible disruption (isolation of a host with active sessions, reboots that drop sessions). The defender does not see this reward decomposition; it sees only the scalar.

\subsection{Testbed}\label{sec:testbed-config}

\begin{figure}
\centering
\includegraphics[width=0.90\columnwidth]{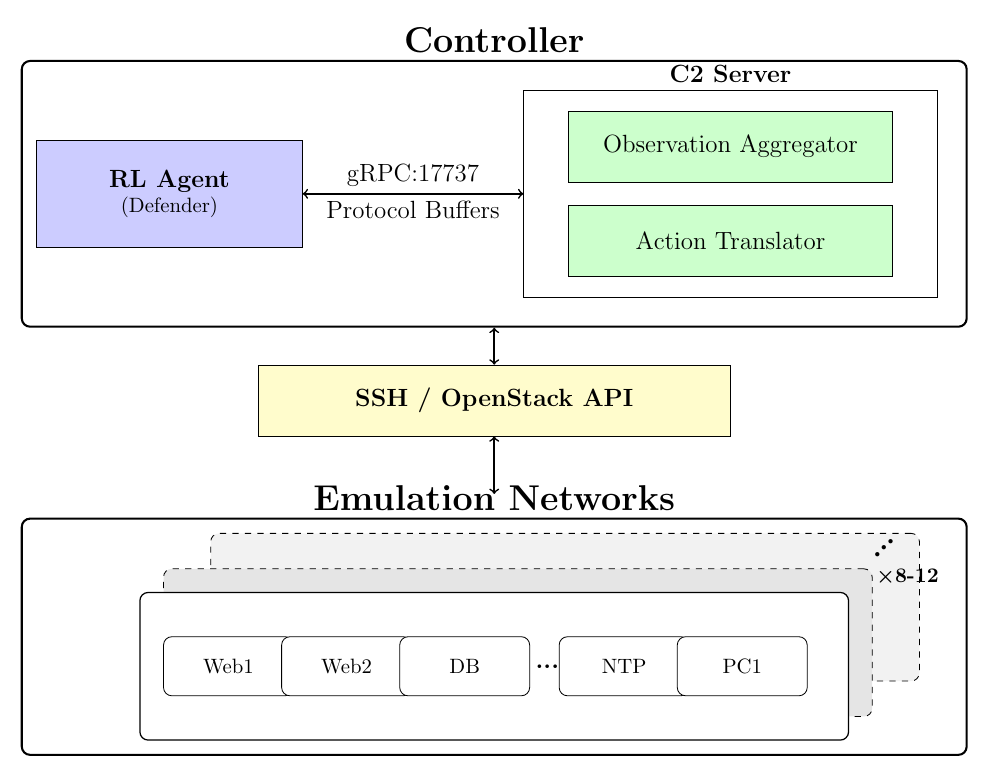}
\caption{Agent-environment interface. The C2 server provides bidirectional translation: abstract defensive tactics become concrete network operations (downward), and distributed sensor data becomes structured observations (upward). Communication occurs over gRPC with Protocol Buffers.}
\label{fig:c2}
\end{figure}

The testbed is an eight-host enterprise emulation deployed on an OpenStack cluster. Every host runs Ubuntu Server 24.04 LTS on a medium VM flavor (2 vCPU, 4\,GiB RAM, 16\,GiB storage). The network is partitioned into three subnets connected by OpenStack security-group rules. The Public Subnet holds two internet-facing WordPress servers (PublicWeb1, PublicWeb2). Private Subnet A holds an NTP server and a database server, reachable only from the public subnet. Private Subnet B holds an internal web server and three user workstations (PC1, PC2, PC3). A separate Controller Subnet, unreachable from any attacker-controlled host, hosts the defender process and the gRPC-based command-and-control server that bridges policy decisions to OpenStack API calls and SSH-driven host operations (Figure~\ref{fig:c2}).

Four classes of vulnerability are deployed across the network (Table~\ref{tab:vuln}). The WordPress Backup Migration flaw (CVSS 9.8) is deployed on both public servers and serves as the canonical entry vector. Weak SSH credentials are configured on the NTP, database, and internal web servers. A weak \texttt{sudo} version is installed on the NTP, database, and internal web servers. A world-writable shell script executed by cron as root is planted on both public servers. Together, these support the kill-chain actions specified in Section~\ref{sec:threat-model}: web exploitation gives initial foothold, weak credentials enable lateral movement, weak \texttt{sudo} and the cron script enable escalation.

\begin{table}
\centering
\caption{CVE-class vulnerabilities deployed in the testbed.}
\label{tab:vuln}
\small
\begin{tabular}{p{0.46\columnwidth} p{0.18\columnwidth} p{0.20\columnwidth}}
\toprule
\textbf{Vulnerability} & \textbf{Severity (CVSS)} & \textbf{Hosts} \\
\midrule
WordPress Backup Migration & 9.8 & PublicWeb1, PublicWeb2 \\
Weak SSH credentials & N/A & NTP, DB, WEB \\
Weak \texttt{sudo} version & N/A & NTP, DB, WEB \\
World-writable cron-as-root & N/A & PublicWeb1, PublicWeb2 \\
\bottomrule
\end{tabular}
\end{table}

User activity is generated by the GHOSTS framework \cite{updyke2018ghosts}. GHOSTS personas perform browsing (5--10 HTTP requests per session over 10--20\,s), SSH administration (20--30\,s sessions), database queries (sub-second to several seconds), and file transfers (30--60\,s), with staggered start times and idle gaps between actions. The user behavior is scripted but human-correlated: bursts and pauses follow distributions calibrated to enterprise traces rather than Poisson noise. Two profiles are used in evaluation. U-rand is memoryless: at every step each user samples an action independently. U-burst follows ON/OFF dynamics with heavy-tail ON durations, which produces extended periods of concentrated activity that overlap attacker actions in the alert stream. User actions fail and produce alerts when their target host is isolated or rebooted, which propagates to the defender's reward. The C2 interface aggregates alerts and timestamped events into the per-host observations (Figure~\ref{fig:c2}).

\textbf{Parallel network environments.} The environment is instantiated as multiple parallel network replicas, each reproducing the full topology with identical machine images, configuration baselines, and vulnerability sets. Hostnames, subnet identifiers, and security objects are namespaced per replica to avoid identifier collisions. Each replica is fully isolated at the cloud-network layer: separate virtual private clouds, non-overlapping CIDR blocks, and replica-scoped security-group namespaces prevent cross-traffic. The C2 service maintains a dedicated gRPC endpoint per replica, handling action translation and sensor aggregation independently for each instance. Telemetry streams, logs, and snapshots are partitioned by replica, and lifecycle operations such as snapshotting, reset, and tear down are executed atomically within a single replica context to preserve reproducibility and fault isolation.

The boundary between what is real and what is emulated is as follows. Real: the OpenStack substrate, the deployed services (WordPress, SSH, NTP, PostgreSQL, cron), the CVE-bound exploit code paths, the wall-clock 5-second polling, and the operational consequences of defender actions on running VMs (an isolated host actually drops user traffic). Emulated: the user population (GHOSTS personas, scripted), the attacker population (two fixed-class profiles, scripted), and the CVE catalog (finite, named in Table~\ref{tab:vuln}). The testbed sits one rung above CAGE-style abstract gyms and CybORG-AWS emulators on the realism ladder, and below larger OpenStack testbeds such as MHBench in scale (8 hosts versus 20--50). The realism is depth-over-breadth: fewer hosts, more end-to-end fidelity per host.

\section{Approach}\label{sec:approach}

\subsection{Preliminaries}

FOE-Dreamer trains a defender policy directly in the operational network of Section~\ref{sec:testbed}. The defender observes a host-level telemetry vector every five seconds, selects one of five tactical action categories per host, and receives a scalar reward summarizing containment quality and user disruption. We cast this loop as a partially observable Markov decision process and introduce the latent factoring on which the rest of the section builds.

The defender's problem is a POMDP $\langle \mathcal{S}, \mathcal{A}, \mathcal{O}, \mathcal{T}, \mathcal{R}, \gamma \rangle$. The action space $\mathcal{A}$ and reward function $\mathcal{R}$ are as specified in Section~\ref{sec:defender}; $\gamma \in (0, 1)$ is the discount factor. The remaining components are:

\begin{itemize}
\item $\mathcal{S}$, the hidden network state. Per host it includes attacker compromise level, privilege escalation, attacker progress along the kill chain, exogenous user activity, and service availability. The defender never observes $\mathcal{S}$ directly.
\item $\mathcal{O}$, the per-host observation vector $o_t \in \mathbb{R}^{N \times F}$: alert counts, honeypot status, host availability, and trap indicators across the $N$ hosts.
\item $\mathcal{T}$, the transition kernel. Transitions arise from the joint effect of the defender action, the unobserved attacker action, and exogenous user activity emitted by GHOSTS \cite{updyke2018ghosts}. The kernel is not analytically known; it is the runtime behavior of the deployed software.
\end{itemize}

The defender selects an action $a_t \in \mathcal{A}$ from $o_t$ and the recurrent belief carried over from previous steps. An episode lasts 100 polling intervals (\textasciitilde 8 minutes wall-clock).

The world model factors the latent belief into three components, each capturing a distinct source of variation in $\mathcal{T}$ and $\mathcal{R}$. The controllable latent $z_t$ carries host states whose evolution is action-conditioned: isolation status, deception deployment, observable reboots, and the alert dynamics that follow defender intervention. The exogenous latent $u_t$ is intended to capture background user activity, service load, and periodic maintenance, subject to the assumptions below. The opponent embedding $\tau$ summarizes recent attacker behavior, inferred only from defender observations and the defender's own actions.

Separate posterior and transition heads encode the intended factor boundary; they do not establish independence of the underlying processes. The opponent embedding $\tau$ is produced by a separate recurrent encoder over a sliding window of recent defender observations and actions. The factoring is what the experiments in Section~\ref{sec:disentanglement} probe.

\begin{figure}
\centering
\includegraphics[width=0.90\columnwidth]{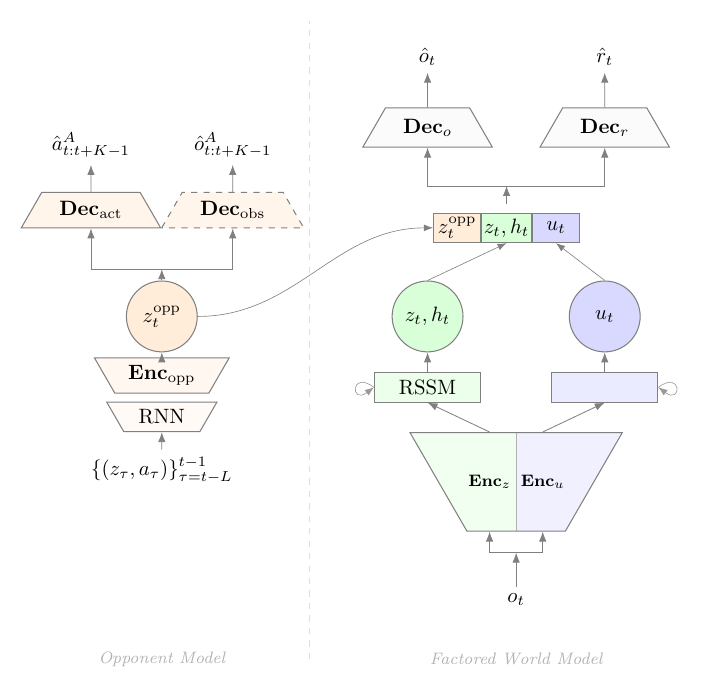}
\caption{FOE-Dreamer architecture overview.}
\label{fig:1}
\end{figure}

Figure~\ref{fig:1} shows the four core components: (i) an observation encoder $f_\theta$, (ii) the factored world model with three branches producing $z_t$, $u_t$, and $\tau$, (iii) the actor-critic policy that conditions on $(z_t, \tau)$, and (iv) the reward predictor that conditions on all three latents. Training proceeds in iterations of three steps: collect $E$ episodes of real interaction in the testbed, fit the world model and opponent encoder on the replay buffer, and update the defender policy via imagined rollouts in latent space. Sections~\ref{sec:wm}--\ref{sec:loop} develop the world model, opponent model, and training loop in turn.

\subsection{Factored World Model}\label{sec:wm}

A single recurrent latent would absorb three qualitatively different signals at once: defender-driven host state changes, exogenous user-driven traffic, and attacker activity. This conflation hurts credit assignment. If an alert spike coincides with both a user logging in and a defender Isolate action, an unfactored latent has no architectural reason to attribute the spike to one or the other. The variance of the policy gradient inflates accordingly. FOE-Dreamer therefore separates the controllable from the uncontrollable so that policy gradients update only the latent the policy can steer.

Let $o_t$ denote the observation at step $t$ and $a_{t-1}$ the defender's previous action. The observation encoder produces a per-step feature
\begin{equation}
h_t = f_{\theta}(o_t).
\end{equation}
Two posterior heads split the encoded signal into a controllable and an exogenous component. The controllable posterior conditions on the previous defender action; the exogenous posterior does not:
\begin{equation}
q_\phi(z_t \mid h_t, s_{t-1}, a_{t-1}), \qquad q_\psi(u_t \mid h_t, s_{t-1}),
\end{equation}
where $s_{t-1} = (z_{t-1}, u_{t-1})$ is the previous recurrent state. The asymmetry is the architectural commitment of the factoring: the action gate enters $z_t$ alone.

\textbf{Factoring assumptions.} Separation assumes (i) action-informative observations, (ii) action-independent user dynamics, (iii) attacker-only training labels, since action and reward conditioning cannot distinguish attacker from user variation, and (iv) additive, action-independent user reward. Containment violates (ii) and (iv): isolation and reboots terminate sessions, coupling disruption costs with user state. Factoring is an inductive bias; reward prediction retains user context through $u_t$.

The two transition models mirror the posteriors. The controllable transition is action-conditioned, the exogenous transition is not:
\begin{equation}
p_\phi(z_t \mid s_{t-1}, a_{t-1}), \qquad p_\psi(u_t \mid u_{t-1}).
\end{equation}
The decoder reconstructs the observation from both factors,
\begin{equation}
\hat{o}_t = g_\eta(z_t, u_t),
\end{equation}
and the reward predictor consumes all three latents because the reward depends on the attacker's behavior as well as on the defender's and the user's:
\begin{equation}
\hat{r}_t = h_\zeta(z_t, u_t, \tau).
\end{equation}

\textbf{Training objective.} The world model is trained with a Dreamer-family evidence lower bound \cite{hafner2023dreamerv3} consisting of an observation reconstruction term, a reward reconstruction term, and a per-factor Kullback--Leibler (KL) divergence between posterior and prior:
\begin{equation}
\begin{aligned}
\mathcal{L}_{\text{WM}} = \, & \mathbb{E}_q\!\Big[ -\log p(o_t \mid z_t, u_t) - \log p(r_t \mid z_t, u_t, \tau) \Big] \\
& + \beta_z \, \mathrm{KL}\big[q_\phi \,\|\, p_\phi\big] + \beta_u \, \mathrm{KL}\big[q_\psi \,\|\, p_\psi\big].
\end{aligned}
\end{equation}
The two KL coefficients $\beta_z$ and $\beta_u$ are tuned independently. Separate transition heads provide an inductive bias, but the controllable posterior and transition still depend on the previous joint state $s_{t-1}=(z_{t-1},u_{t-1})$; the factors are not strictly isolated. Empirically, the action-conditioned $z_t$ tends to predict defender-induced effects while the action-free $u_t$ tends to predict user activity; Section~\ref{sec:disentanglement} reports the diagnostic.

\subsection{Opponent Model}\label{sec:opponent}

The defender cannot directly observe attacker actions or intent, but attacker behavior leaves signatures in the defender's telemetry. An Avoidant attacker withdraws when a honeypot triggers; an Enticed attacker presses on against the same signal. These signatures unfold over multiple observations. An opponent model trained to predict the attacker's next action and target host lets the policy condition on inferred behavior rather than wait for the attacker to commit a revealing action. The expected effect is earlier containment against probe-and-withdraw adversaries \cite{fergusonwalter2019tularosa}.

The recurrent opponent encoder reads $K$ defender observation/action pairs, available at training and deployment:
\begin{equation}
\tau_t = f_\xi\!\left(\{(o_j,a_{j-1})\}_{j=t-K+1}^{t}\right).
\end{equation}
The embedding $\tau_t$ feeds the world model's reward predictor and the policy; it does not enter the observation decoder $g_\eta$. A decoder with parameters $\omega$ has two categorical prediction heads, one for the attacker's next action and one for its next target host. There is no attacker-class classifier and no opponent-observation decoder. The encoder and prediction heads are trained through the objective in Section~\ref{sec:loop}.

\textbf{Privileged-information training.} During training the testbed exposes the attacker's action and target-host sequence as prediction targets. Neither the policy, the world model, nor the opponent encoder receives this sequence as an input. The two decoder heads are supervised by cross-entropy losses on the next attacker action and target host; attacker-class labels are not the training objective. At deployment only the encoder $f_\xi$ is needed to compute $\tau_t$ from defender-observable history, without a ground-truth attacker feed. Class separation in the embedding visualization is a diagnostic, not a classifier used for training.

\subsection{Live Training Loop}\label{sec:loop}

Let $\theta, \phi, \psi, \eta, \zeta$ denote the world-model parameters (encoder, transitions, decoder, reward predictor), $\xi$ the opponent encoder, $\omega$ its two prediction heads, $\pi$ the defender actor, and $v$ the critic. One training iteration runs as follows:

\begin{enumerate}
\item \textbf{Collect.} Run the current policy in $E$ parallel testbed instances for one episode each. Log $(o_t, a_t, r_t)$ trajectories and the attacker action and target-host sequence used as training targets. Append to the replay buffer.
\item \textbf{Fit.} Sample minibatches of trajectories from the buffer and take gradient steps on $\mathcal{L}_{\text{WM}}$ to update $(\theta, \phi, \psi, \eta, \zeta)$ and on the two opponent-prediction losses to update $(\xi,\omega)$.
\item \textbf{Imagine.} Sample initial latents from the buffer, then unroll the learned world model forward for $H$ steps using the current actor; recompute $\tau$ at each imagined step from the imagined observation and defender-action window.
\item \textbf{Optimize.} Update $\pi$ and $v$ via actor-critic on the imagined rollouts. No real-environment steps are taken in this phase.
\end{enumerate}

Iterations repeat until the wall-clock training budget is exhausted. Real interaction is reserved for the Collect phase; the bulk of policy updates happens in latent imagination, keeping the 3-day budget tractable.

\textbf{Joint objective.} The combined loss is
\begin{equation}
\mathcal{L}_{\text{total}} = \mathcal{L}_{\text{WM}} + \alpha\,\mathcal{L}_{\text{policy}} + \beta\,\mathcal{L}_{\text{opponent}},
\end{equation}
where $\mathcal{L}_{\text{policy}}$ is the actor-critic loss over imagined rollouts and $\mathcal{L}_{\text{opponent}}$ is the sum of the next-action and next-target cross-entropy losses. The weights $\alpha$ and $\beta$ are set by the hyperparameter search in Appendix~\ref{app:hyperparameters}. Table~\ref{tab:loss_supervision} identifies the directly supervised modules for each objective. The privileged attacker targets supervise the opponent-prediction branch only; they are not policy inputs.

\begin{table}[t]
\centering
\caption{Objectives and directly supervised parameter blocks.}
\label{tab:loss_supervision}
\small
\begin{tabular}{@{}p{0.17\columnwidth}p{0.28\columnwidth}p{0.44\columnwidth}@{}}
\toprule
Objective & Targets & Parameter blocks \\
\midrule
World model & $o_t$, $r_t$; latent priors & $(\theta,\phi,\psi,\eta,\zeta)$ \\
Opponent & Next attacker action and target & Encoder $\xi$, heads $\omega$ \\
Actor-critic & Imagined returns & Actor $\pi$, critic $v$ \\
\bottomrule
\end{tabular}
\end{table}

\section{Evaluation}\label{sec:eval}

The evaluation answers three questions in turn. First, does FOE-Dreamer learn a better defender than model-free baselines under matched compute on the live testbed? Section~\ref{sec:sample} addresses this with the headline comparison against Rainbow, IQN, and three non-learning reference policies. Second, do the two architectural contributions actually carry the gain? Section~\ref{sec:ablation} ablates the factored world model and the opponent model. Third, does the factoring do what it is designed to do, or is the gain a side effect of something else? Section~\ref{sec:disentanglement} probes the learned latents directly. Section~\ref{sec:deployment} closes with notes on the operational behavior of the system on the testbed.

\subsection{Setup}\label{sec:setup}

The setup is held constant across the questions above. Learning-based methods are trained on a single GPU for 3 days of wall-clock time per configuration. Each run uses 8 parallel environment instances on the OpenStack testbed of Section~\ref{sec:testbed-config}, with 5 random seeds per configuration; we report mean and standard deviation across seeds. Raw per-seed episode losses are included with the artifact bundle so that bootstrap confidence intervals can be computed independently.

An episode lasts 100 timesteps with a polling interval of 5 seconds, for roughly 8 minutes of wall-clock time per episode. Defender observations and actions are issued once per polling step. The batch $E=8$ denotes episodes per iteration, one per environment instance; collection and learning repeat throughout training (Section~\ref{sec:loop}).

\textbf{Baselines.} The learned baselines are two standard model-free distributional RL methods used in prior autonomous cyber-defense work: Rainbow \cite{hessel2018rainbow} and IQN \cite{dabney2018iqn}. Both train on the same observation stream, the same action space, the same compute budget, and the same seeds as FOE-Dreamer. FOE-Dreamer additionally uses attacker action/target labels to train opponent prediction, never as policy inputs; deployed policies use defender-observable history.

Three non-learning reference policies bracket the achievable range. \textit{Sleep} always selects the Null action; it never intervenes and gives a no-defense lower bound. \textit{Random} samples uniformly over the five-element defender action space at each polling step; it shows what undirected activity buys against each attacker. \textit{Expert-Inspired Heuristic} is a rule-based defender that, at the start of each episode, deploys a Cowrie honeypot on one public-facing server and plants a honeyfile on a randomly chosen internal server; during the episode it isolates any host that triggers a decoy or raises an escalation alert and reboots an isolated host once if no further alerts arrive. The heuristic encodes the standard cyber-deception playbook \cite{fergusonwalter2019tularosa,fergusonwalter2021decoy} as a fixed policy and serves as the operational floor that a learned defender is expected to exceed.

\textbf{Attacker and user profiles.} The two scripted attacker profiles are Avoidant (cautious, probe-first, withdraws on positive deception indicators) and Enticed (opportunistic, rapid-exploitation), as defined in Section~\ref{sec:threat-model}. Background user activity is generated by GHOSTS under U-rand (memoryless, independent action per step) and U-burst (ON/OFF phases with heavy-tail ON durations), as defined in Section~\ref{sec:testbed-config}.

\textbf{Metric.} The primary metric is \textbf{episode loss}, the negative reward sum over the 100-step episode. Lower (more negative) is worse. Variability is reported as the standard deviation across the five training seeds.

\subsection{Sample Efficiency}\label{sec:sample}

\begin{finding}
\textbf{Finding 1:} FOE-Dreamer reduces episode loss against both attacker profiles by roughly $2\times$ relative to Rainbow and IQN within the same 3-day training budget.
\end{finding}

Table~\ref{tab:headline} reports final episode losses after training. FOE-Dreamer surpasses every reference and learned baseline against both attacker profiles. Among learned baselines, FOE-Dreamer's $-48$ is roughly half the magnitude of Rainbow's $-102$ and well below IQN's $-118$ against Avoidant; against Enticed, FOE-Dreamer's $-38$ is roughly half of Rainbow's $-76$ and IQN's $-84$. The gap exceeds one standard deviation in every cell, so the ordering is stable across seeds. FOE-Dreamer also surpasses the Expert-Inspired heuristic in both columns: under three days of live training, the learned defender exceeds the rule-based operational floor.

\begin{table}
\centering
\caption{Final episode loss (mean $\pm$ std over 5 seeds) after 3 days of live training. Lower is worse. Sleep, Random, and Expert-Inspired Heuristic are reference baselines; Rainbow and IQN are learned baselines.}
\label{tab:headline}
\small
\begin{tabular}{lcc}
\toprule
Method & Avoidant & Enticed \\
\midrule
\textbf{FOE-Dreamer} & $\mathbf{-48 \pm 7}$ & $\mathbf{-38 \pm 5}$ \\
Rainbow & $-102 \pm 20$ & $-76 \pm 17$ \\
IQN & $-118 \pm 26$ & $-84 \pm 21$ \\
Expert & $-138 \pm 18$ & $-88 \pm 15$ \\
Random & $-152 \pm 28$ & $-242 \pm 38$ \\
Sleep & $-178 \pm 24$ & $-285 \pm 32$ \\
\bottomrule
\end{tabular}
\end{table}

\begin{figure*}[t]
\centering
\includegraphics[width=0.80\textwidth]{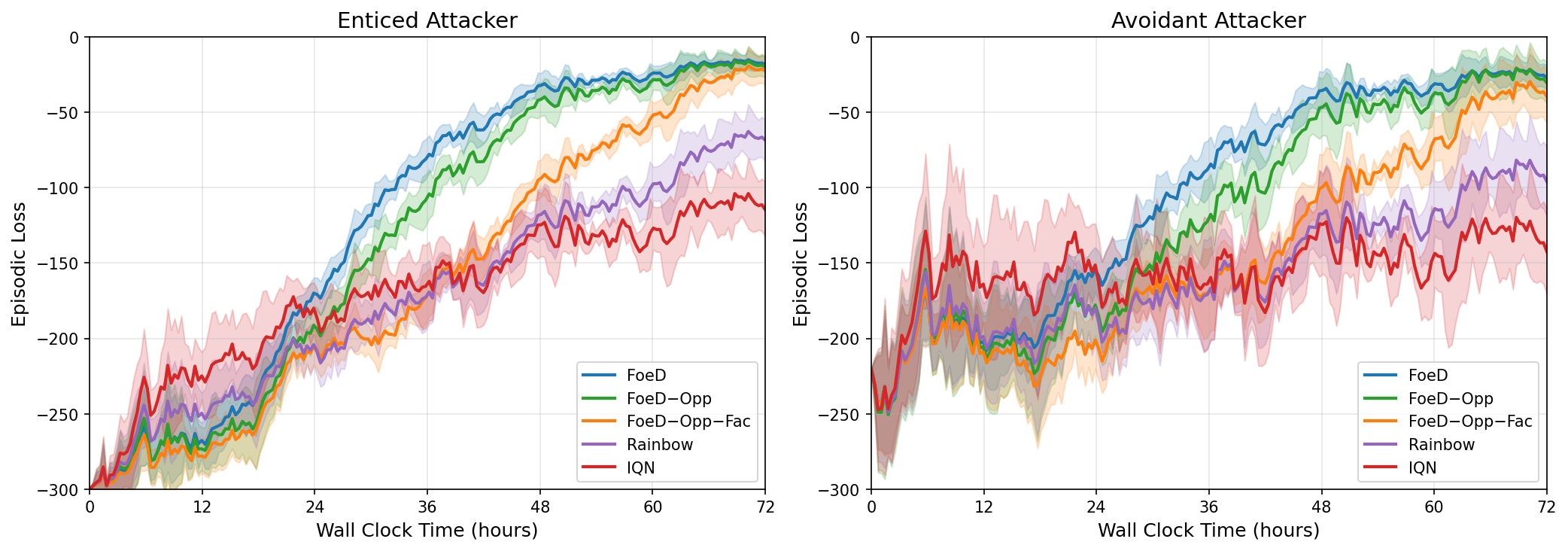}
\caption{Episode loss vs.\ training day across baselines and FOE-Dreamer, under both attacker profiles.}
\label{fig:2}
\end{figure*}

The headline numbers compress the trajectory; Figure~\ref{fig:2} shows the full learning curves across the 3-day budget. Rainbow and IQN flatten before day 3 against Avoidant, while FOE-Dreamer continues to improve; against Enticed, all methods improve, but FOE-Dreamer improves faster and achieves better final performance. The variance pattern is consistent across configurations: FOE-Dreamer has the tightest seed-to-seed bands, the Full$-$Factor variant somewhat wider, the Full$-$Factor$-$Opponent variant wider still, and the model-free baselines widest. Opponent inference resolves the ambiguity created by probe-and-withdraw behavior, and exogenous factoring routes user-driven variability out of the controllable latent; the two effects appear first in the variance bands and then in the means.

\subsection{Ablations: Component Contributions}\label{sec:ablation}

The headline numbers leave open which architectural component is doing the work. To isolate the contributions of the factored world model and the opponent model, we compare three configurations:

\begin{itemize}
\item \textbf{Full}: factored world model + opponent model (FOE-Dreamer).
\item \textbf{Full$-$Factor}: opponent model retained, but a single $z_t$ replaces the $z_t / u_t$ split.
\item \textbf{Full$-$Factor$-$Opponent}: neither factoring nor opponent modeling. Single-latent world model only.
\end{itemize}

\begin{table}
\centering
\caption{Ablation matrix. Final episode loss (mean $\pm$ std over 5 seeds) after 3 days. Lower is worse. Full$-$Factor$-$Opponent is the single-latent Dreamer baseline.}
\label{tab:ablation}
\begin{tabular}{lcc}
\toprule
Configuration & Avoidant & Enticed \\
\midrule
Full (FOE-Dreamer) & $-48 \pm 7$ & $-38 \pm 5$ \\
Full$-$Factor & $-62 \pm 12$ & $-48 \pm 11$ \\
Full$-$Factor$-$Opponent & $-78 \pm 16$ & $-58 \pm 14$ \\
\bottomrule
\end{tabular}
\end{table}

\begin{finding}
\textbf{Finding 2:} Both factoring and opponent modeling contribute meaningfully. Opponent modeling matters most against the Avoidant attacker; factoring matters across both attacker profiles.
\end{finding}

Removing exogenous factoring (Full to Full$-$Factor) costs about 14 points of episode loss against Avoidant and 10 points against Enticed; the standard deviation also widens. Removing the opponent model on top (Full$-$Factor to Full$-$Factor$-$Opponent) costs another 16 points against Avoidant and 10 against Enticed.

Two patterns are worth flagging. \textbf{Opponent modeling matters most against Avoidant.} The Avoidant column drops by 16 points ($-62$ to $-78$) when the opponent head is removed, versus 10 points on Enticed. Avoidant's probe-and-withdraw behavior creates extended ambiguity windows that the opponent model resolves; without it, the policy commits late or wrong. \textbf{Exogenous factoring matters across both profiles.} Both columns lose meaningful ground ($-48$ to $-62$ and $-38$ to $-48$) when $u_t$ is folded back into $z_t$, because user activity then leaks into the controllable latent and inflates value-target variance during planning.

The ablation is nested, not a complete $2\times2$ design: Factor+No-Opponent was considered but excluded because opponent supervision supports attacker/user separation. The comparisons therefore do not establish full factorial main or interaction effects.

The variance pattern follows the same direction. Across all three rows, the Full configuration has the tightest standard deviation ($\pm 7$, $\pm 5$); each ablation widens the band. Factoring removes a noise channel and the opponent model removes a hidden-mode source of variance, and seed-to-seed reproducibility improves when both are present.

\subsection{Latent Disentanglement Validation}\label{sec:disentanglement}

The ablations show that the components help. They do not show whether the components do what we claim or whether they help by some other mechanism. This subsection probes the learned latents directly: whether $u_t$ actually captures user-driven variation, whether $z_t$ stays focused on defender-driven dynamics, and whether the opponent embedding $\tau$ separates attacker types in a way that supports action prediction.

\textbf{Exogenous latent captures user activity.} We define ground-truth user activity as the count of user-originated observable events in the last five steps (LocalWork failures, LocalWork-induced false alerts, AccessService-induced false network alerts), then fit ridge regressions from $u_t$ and from $z_t$ to that target on held-out trajectories.

\begin{table}
\centering
\caption{Linear-decoding $R^2$ for user activity from each latent. Higher means stronger linear relationship.}
\label{tab:r2-user}
\begin{tabular}{lcc}
\toprule
Input latent & $R^2$ (U-rand) & $R^2$ (U-burst) \\
\midrule
$u_t$ & 0.72 & 0.79 \\
$z_t$ & 0.14 & 0.18 \\
\bottomrule
\end{tabular}
\end{table}

\begin{finding}
\textbf{Finding 3:} User activity is decoded much more accurately from the exogenous latent than from the controllable latent ($R^2$ 0.72--0.79 vs.\ 0.14--0.18). The factored world model routes user-driven variation away from the component the policy can steer, as intended.
\end{finding}

$u_t$ predicts user activity well (0.72 under U-rand, 0.79 under U-burst). $z_t$ predicts user activity poorly (0.14, 0.18). This is the asymmetry the architecture is supposed to enforce: user-driven variation should land in $u_t$ and not contaminate $z_t$.

\textbf{Controllable latent captures defender actions.} We fit multinomial logistic classifiers from each latent to the defender's discrete action $a_t$ on held-out trajectories. Random-baseline accuracy is 0.17 over six action categories.

\begin{table}
\centering
\caption{Defender-action prediction accuracy from each latent.}
\label{tab:action-pred}
\begin{tabular}{lc}
\toprule
Input latent & Accuracy \\
\midrule
$u_t$ & 0.21 \\
$z_t$ & 0.71 \\
\bottomrule
\end{tabular}
\end{table}

\begin{finding}
\textbf{Finding 4:} Defender actions are decoded much more accurately from the controllable latent than from the exogenous latent (0.71 vs.\ 0.21 accuracy, against a 0.17 random baseline). The controllable component captures action-driven dynamics; the exogenous component does not.
\end{finding}

$z_t$ recovers the defender's action with 0.71 accuracy. $u_t$ is at near-chance (0.21 vs.\ 0.17 random). The two tables together give a testable check on the factoring: $u_t$ tracks users and ignores the defender, $z_t$ tracks the defender and ignores the user.

\begin{figure}
\centering
\includegraphics[width=0.90\columnwidth]{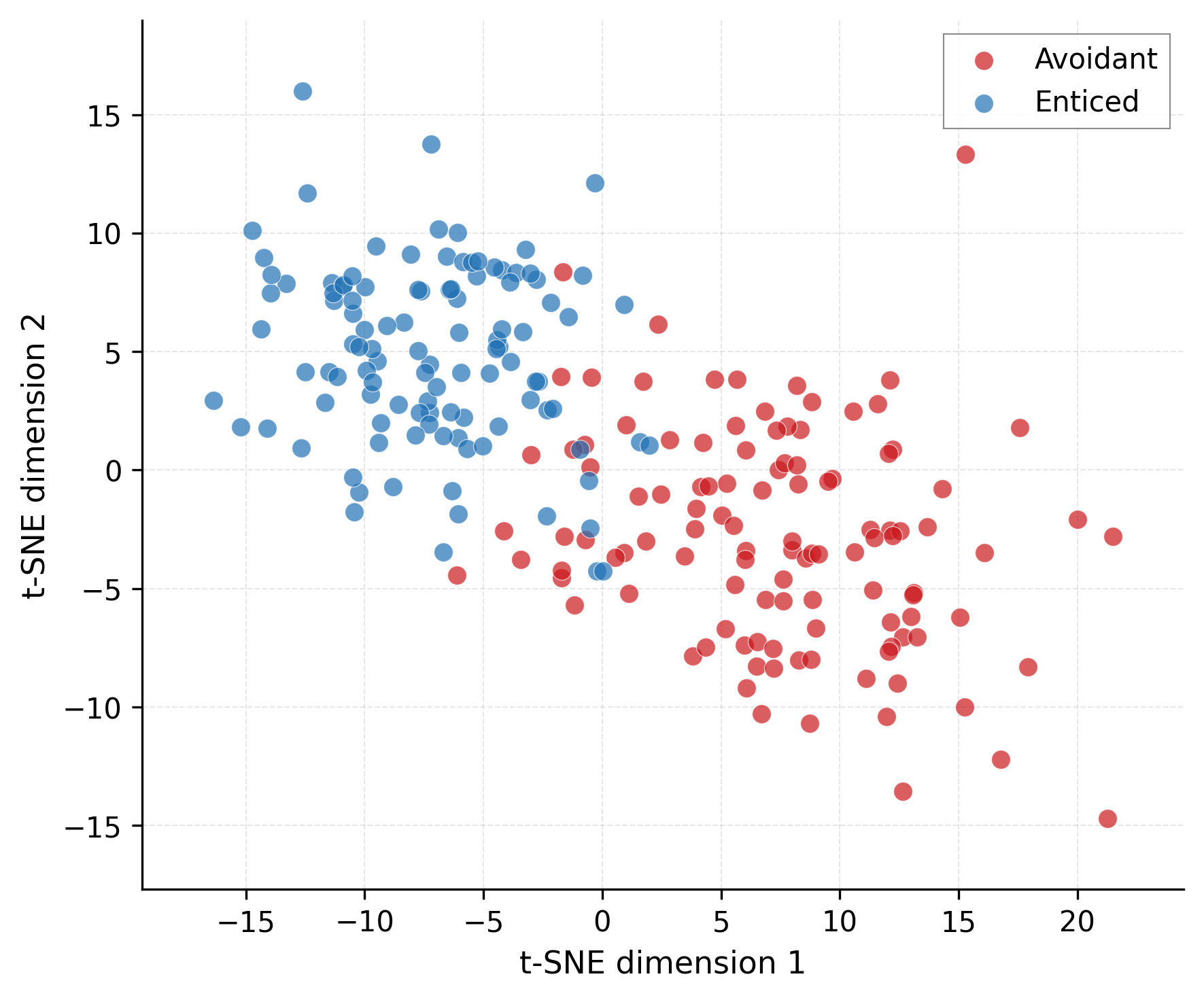}
\caption{t-SNE projection of the opponent embedding $\tau$ at a fixed step in each episode, colored by attacker profile. The two profiles occupy clearly separated regions of the embedding space.}
\label{fig:opp-embed}
\end{figure}

\begin{figure}
\centering
\includegraphics[width=0.90\columnwidth]{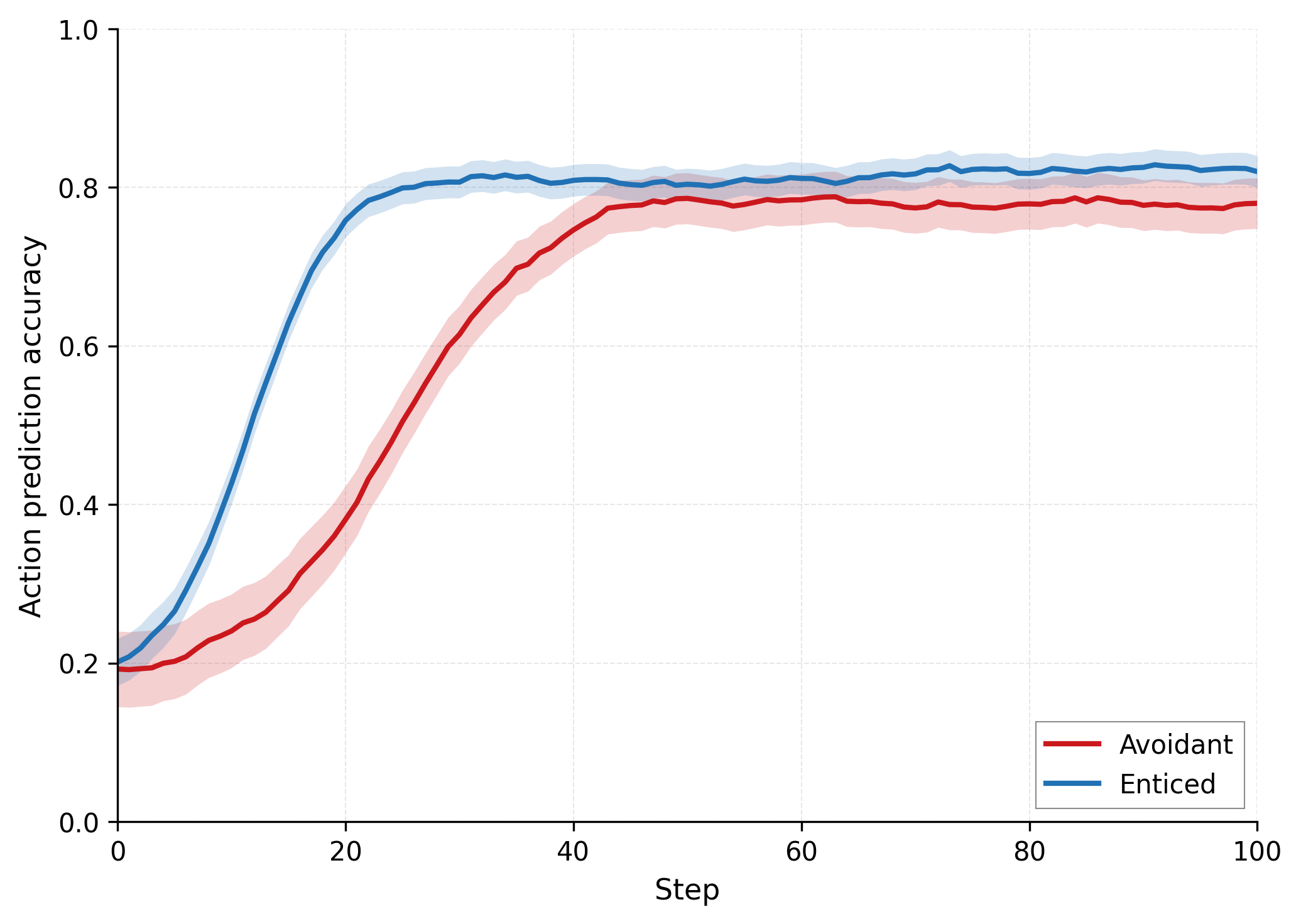}
\caption{Action-prediction accuracy of the opponent encoder over training, by attacker profile. Accuracy rises monotonically as the encoder learns to summarize defender-observable signals.}
\label{fig:opp-acc}
\end{figure}

\textbf{Opponent embedding separates attacker profiles.} Figures~\ref{fig:opp-embed} and~\ref{fig:opp-acc} show a t-SNE projection of the opponent embedding $\tau$ at a fixed step in each episode, colored by attacker profile. The two profiles occupy clearly separated regions of the embedding space, which is what the policy needs to condition on attacker type. Action-prediction accuracy of $\tau$ rises monotonically over training, indicating that the opponent encoder learns increasingly informative summaries from defender-observable signals alone.

Together, the diagnostics indicate that the architectural factoring is interpretable, testable, and consistent with its intended role.

\subsection{Deployment Notes}\label{sec:deployment}

\begin{finding}
\textbf{Finding 5:} Running FOE-Dreamer on a live OpenStack testbed surfaces three operational realities --- action-latency asymmetry, user-emulation drift, and snapshot-restore cost --- that future deployments will need to anticipate.
\end{finding}

The numbers above were obtained under specific operational conditions on the OpenStack testbed, and three of those conditions surfaced repeatedly during the 3-day budget. First, \textbf{action latency varies}: SSH-driven Disrupt actions complete in 2--6 seconds depending on host load, while OpenStack security-group changes for Isolate complete in under a second; the policy is exposed to this asymmetry through the polling interval and learns to prefer Isolate when both options are otherwise equivalent. Second, \textbf{user emulation drift}: GHOSTS personas occasionally desynchronise between testbed instances when one VM lags, producing brief windows where the same nominal user profile generates different observable activity across replicas; this contributes to the variance reported in Table~\ref{tab:headline}. Third, \textbf{snapshot-and-restore cost}: episode resets via Eradicate require restoring all eight VMs from snapshot, taking 35--90 seconds. The Collect phase dominates wall-clock cost overall: each episode takes roughly 8 minutes, and the 8 parallel testbed instances execute on the same OpenStack tenant, so background load on the cluster is high during training. These observations are not empirical claims; they describe the conditions under which the reported numbers were obtained and the operational regime a future deployment would have to accommodate.

\section{Discussion and Limitations}

\textbf{Emulation-to-deployment gap.} FOE-Dreamer trains in a higher-fidelity emulation, not on a production network. The OpenStack substrate, real CVE replay, and GHOSTS-generated user activity narrow the gap relative to abstract simulators, but transferring a trained policy to a production network with real attackers and real users remains future work. The 3-day training budget reported here does not imply equivalent sample requirements in production.

\textbf{Attacker coverage.} Evaluation is limited to the trained, scripted Avoidant and Enticed classes and their within-class variation. Zero-shot transfer to unseen attacker classes and adaptive learning adversaries is not demonstrated. The profiles were chosen for a defender with deception actions; they do not establish coverage of other behavioral differences.

\textbf{Known failure cases.} Four recurring failure modes appear during evaluation. (i) \textit{Deception budget exhausted early.} Against the Enticed attacker the policy occasionally deploys both available deception artifacts within the first 30 timesteps, leaving none in reserve when the attacker pivots to a second host; recovery requires a Disrupt that disrupts users. (ii) \textit{Oscillation under Avoidant probing.} Against the Avoidant attacker the policy sometimes alternates between Isolate and a Restore-by-reboot loop while the opponent encoder is still uncertain about attacker class, producing unnecessary user disruption before $\tau$ stabilizes. (iii) \textit{Exogenous-prediction degradation under U-burst.} When GHOSTS shifts from low-rate to burst regime mid-episode, $u_t$ briefly tracks the controllable latent, which inflates the value-target variance until the burst phase ends. (iv) \textit{Late detection on cron-escalation.} The world-writable cron path produces no observable alert until the next cron tick fires, so any policy --- ours included --- has a one-cron-period blind window after the relevant write.

\textbf{Deferred analyses.} A hyperparameter sensitivity sweep and a quantitative prevalence count of the failure modes above are deferred to follow-on work; both require additional training and evaluation cycles beyond the current compute budget.

\textbf{Scaling.} The testbed is small relative to enterprise networks. Scaling FOE-Dreamer to 50+ host networks is a deployment question not addressed here. The factoring and opponent components are expected to remain useful at larger scale, because the underlying problems (exogenous user noise, attacker-type uncertainty) only intensify with more hosts.

\section{Ethical Considerations}

All CVEs deployed in the testbed (WordPress Backup Migration CVSS 9.8, weak SSH credentials, weak \texttt{sudo}, world-writable cron-as-root) are publicly disclosed with available patches; no new vulnerabilities were discovered. User activity is synthetic, generated by GHOSTS with scripted personas; no human subjects and no PII are involved at any stage. All experiments ran on internally hosted infrastructure with no external network impact. The released artifact at \url{https://anonymous.4open.science/r/FOE-DREAMER-5117} contains the defender-side code needed to reproduce the empirical claims and does not include runnable exploit chains against external infrastructure.

\section{Conclusion}

We presented FOE-Dreamer, a model-based reinforcement learning framework with two technical contributions: a factored world model that separates defender-driven, exogenous, and adversarial dynamics, and an opponent model that infers attacker type from defender-observable signals. Within a 3-day live-training budget on the OpenStack testbed, FOE-Dreamer reduces episode loss against Avoidant and Enticed attackers by roughly $2\times$ relative to Rainbow and IQN, with both architectural components shown to contribute and the latent factoring supported by linear-decoding diagnostics. Sample-efficient RL training in higher-fidelity emulation is therefore feasible within budgets that direct interaction with a deployed network allows.

\bibliographystyle{IEEEtran} \bibliography{refs}

@article{vyas2025acnd,
  author    = {Vyas, Sanyam and Mavroudis, Vasilios and Burnap, Pete},
  title     = {Towards the Deployment of Realistic Autonomous Cyber Network Defence: A Systematic Review},
  journal   = {ACM Computing Surveys},
  year      = {2025},
  volume    = {58},
  number    = {1},
  articleno = {5},
  pages     = {5:1--5:36},
  month     = aug,
  doi       = {10.1145/3729213},
  url       = {https://doi.org/10.1145/3729213}
}

@article{mcfadden2026sok,
  author  = {McFadden, Shae and Foley, Myles and Bates, Elizabeth and Tsingenopoulos, Ilias and Vyas, Sanyam and Mavroudis, Vasilios and Hicks, Chris and Pierazzi, Fabio},
  title   = {{SoK}: The Pitfalls of Deep Reinforcement Learning for Cybersecurity},
  journal = {arXiv preprint arXiv:2602.08690},
  year    = {2026},
  url     = {https://arxiv.org/abs/2602.08690},
  note    = {Systematization of Knowledge}
}

@misc{bauer2026cyber,
  author       = {Bauer, Lujo and Brumley, David and Calandrino, Joseph and Christin, Nicolas and Fanti, Giulia and Gligor, Virgil and Parno, Bryan and Patel, Jignesh and Sekar, Vyas and Sherry, Justine},
  title        = {Creating a Scientific Foundation for Cyber Autonomy},
  howpublished = {CyLab, Carnegie Mellon University, Whitepaper v1.0.0},
  year         = {2026},
  month        = mar,
  note         = {CMU CyLab Cyber Autonomy whitepaper, v1.0.0, March 20, 2026},
  url          = {https://www.cylab.cmu.edu/}
}

@article{salvato2021reality,
  author  = {Salvato, Erica and Fenu, Gianfranco and Medvet, Eric and Pellegrino, Felice Andrea},
  title   = {Crossing the Reality Gap: A Survey on Sim-to-Real Transferability of Robot Controllers in Reinforcement Learning},
  journal = {IEEE Access},
  year    = {2021},
  volume  = {9},
  pages   = {153171--153187},
  doi     = {10.1109/ACCESS.2021.3126658}
}

@inproceedings{ha2018worldmodels,
  author    = {Ha, David and Schmidhuber, J{\"u}rgen},
  title     = {Recurrent World Models Facilitate Policy Evolution},
  booktitle = {Advances in Neural Information Processing Systems (NeurIPS)},
  year      = {2018},
  pages     = {2451--2463},
  note      = {Also available as ``World Models,'' arXiv:1809.01999},
  url       = {https://arxiv.org/abs/1809.01999}
}

@inproceedings{hafner2020dreamer,
  author    = {Hafner, Danijar and Lillicrap, Timothy and Ba, Jimmy and Norouzi, Mohammad},
  title     = {Dream to Control: Learning Behaviors by Latent Imagination},
  booktitle = {International Conference on Learning Representations (ICLR)},
  year      = {2020},
  url       = {https://arxiv.org/abs/1912.01603}
}

@inproceedings{hafner2021dreamerv2,
  author    = {Hafner, Danijar and Lillicrap, Timothy and Norouzi, Mohammad and Ba, Jimmy},
  title     = {Mastering {Atari} with Discrete World Models},
  booktitle = {International Conference on Learning Representations (ICLR)},
  year      = {2021},
  url       = {https://arxiv.org/abs/2010.02193}
}

@article{hafner2023dreamerv3,
  author  = {Hafner, Danijar and Pasukonis, Jurgis and Ba, Jimmy and Lillicrap, Timothy},
  title   = {Mastering Diverse Domains through World Models},
  journal = {arXiv preprint arXiv:2301.04104},
  year    = {2023},
  url     = {https://arxiv.org/abs/2301.04104}
}

@inproceedings{hessel2018rainbow,
  author    = {Hessel, Matteo and Modayil, Joseph and van Hasselt, Hado and Schaul, Tom and Ostrovski, Georg and Dabney, Will and Horgan, Dan and Piot, Bilal and Azar, Mohammad Gheshlaghi and Silver, David},
  title     = {Rainbow: Combining Improvements in Deep Reinforcement Learning},
  booktitle = {Proceedings of the 32nd AAAI Conference on Artificial Intelligence (AAAI)},
  year      = {2018},
  pages     = {3215--3222},
  url       = {https://arxiv.org/abs/1710.02298}
}

@inproceedings{dabney2018iqn,
  author    = {Dabney, Will and Ostrovski, Georg and Silver, David and Munos, R{\'e}mi},
  title     = {Implicit Quantile Networks for Distributional Reinforcement Learning},
  booktitle = {Proceedings of the 35th International Conference on Machine Learning (ICML)},
  year      = {2018},
  pages     = {1096--1105},
  url       = {https://arxiv.org/abs/1806.06923}
}

@article{standen2021cyborg,
  author  = {Standen, Maxwell and Lucas, Martin and Bowman, David and Richer, Toby J. and Kim, Junae and Marriott, Damian},
  title   = {{CybORG}: A Gym for the Development of Autonomous Cyber Agents},
  journal = {arXiv preprint arXiv:2108.09118},
  year    = {2021},
  note    = {IJCAI-21 1st Workshop on Adaptive Cyber Defense (ACD)},
  url     = {https://arxiv.org/abs/2108.09118}
}

@article{kiely2025cage4,
  author  = {Kiely, Mitchell and Ahiskali, Metin and Borde, Etienne and Bowman, Benjamin and Bowman, David and Van Bruggen, Dirk and Cowan, K. C. and Dasgupta, Prithviraj and Devendorf, Erich and Edwards, Ben and Fitts, Alex and Fugate, Sunny and Gabrys, Ryan and Gould, Wayne and Huang, H. Howie and Jacobs, Jules and Kerr, Ryan and King, Isaiah J. and Li, Li and Martinez, Luis and Moir, Christopher and Murphy, Craig and Naish, Olivia and Owens, Claire and Purchase, Miranda and Ridley, Ahmad and Taylor, Adrian and Farmer, Sara and Valentine, William John and Zhang, Yiyi},
  title   = {{CAGE} Challenge 4: A Scalable Multi-Agent Reinforcement Learning Gym for Autonomous Cyber Defence},
  journal = {AI Magazine},
  year    = {2025},
  volume  = {46},
  number  = {3},
  pages   = {e70021},
  doi     = {10.1002/aaai.70021},
  publisher = {Wiley},
  url     = {https://doi.org/10.1002/aaai.70021}
}

@inproceedings{oesch2024cyberwheel,
  author    = {Oesch, Sean and Chaulagain, Amul and Weber, Brian and Dixson, Matthew and Sadovnik, Amir and Roberson, Benjamin and Watson, Cory and Austria, Phillipe},
  title     = {Towards a High Fidelity Training Environment for Autonomous Cyber Defense Agents},
  booktitle = {Proceedings of the 17th Cyber Security Experimentation and Test Workshop (CSET)},
  year      = {2024},
  pages     = {91--99},
  doi       = {10.1145/3675741.3675752}
}

@inproceedings{acuto2023yawningtitan,
  author    = {Acuto, Alberto and Maskell, Simon and Jack, D.},
  title     = {Defending the Unknown: Exploring Reinforcement Learning Agents' Deployment in Realistic, Unseen Networks},
  booktitle = {Proceedings of the Conference on Applied Machine Learning in Information Security (CAMLIS)},
  year      = {2023},
  series    = {CEUR Workshop Proceedings},
  volume    = {3652},
  pages     = {22--35},
  url       = {https://ceur-ws.org/Vol-3652/paper2.pdf}
}

@inproceedings{andrew2022yawningtitan,
  author    = {Andrew, Alex and Spillard, Sam and Collyer, Joshua and Dhir, Neil},
  title     = {Developing Optimal Causal Cyber-Defence Agents via Cyber Security Simulation},
  booktitle = {ICML 2022 Workshop on Machine Learning for Cybersecurity (MLC)},
  year      = {2022},
  note      = {Yawning Titan platform; DSTL release, MIT license, July 2022},
  url       = {https://arxiv.org/abs/2207.12355}
}

@misc{microsoft2021cyberbattlesim,
  author       = {{Microsoft Defender Research Team}},
  title        = {{CyberBattleSim}},
  year         = {2021},
  howpublished = {\url{https://github.com/microsoft/CyberBattleSim}},
  note         = {Software release, announced April 8, 2021}
}

@inproceedings{rigaki2024outofcage,
  author    = {Rigaki, Maria and Luk{\'a}{\v{s}}, Ond{\v{r}}ej and Catania, Carlos A. and Garcia, Sebastian},
  title     = {Out of the Cage: How Stochastic Parrots Win in Cyber Security Environments},
  booktitle = {Proceedings of the 16th International Conference on Agents and Artificial Intelligence (ICAART)},
  year      = {2024},
  pages     = {774--781},
  publisher = {SciTePress},
  doi       = {10.5220/0012391800003636},
  isbn      = {978-989-758-680-4},
  url       = {https://arxiv.org/abs/2308.12086}
}

@article{rigaki2024hackphyr,
  author  = {Rigaki, Maria and Catania, Carlos A. and Garcia, Sebastian},
  title   = {{Hackphyr}: A Local Fine-Tuned {LLM} Agent for Network Security Environments},
  journal = {arXiv preprint arXiv:2409.11276},
  year    = {2024},
  url     = {https://arxiv.org/abs/2409.11276}
}

@article{kouremetis2024mirage,
  author  = {Kouremetis, Michael and Lawrence, Dean and Alford, Ron and Cheuvront, Zoe and Davila, David and Geyer, Benjamin and Haigh, Trevor and Michalak, Ethan and Murphy, Rachel and Russo, Gianpaolo},
  title   = {Mirage: Cyber Deception Against Autonomous Cyber Attacks in Emulation and Simulation},
  journal = {Annals of Telecommunications},
  year    = {2024},
  doi     = {10.1007/s12243-024-01018-4}
}

@article{hammar2023csle,
  author  = {Hammar, Kim and Stadler, Rolf},
  title   = {Learning Near-Optimal Intrusion Responses Against Dynamic Attackers},
  journal = {IEEE Transactions on Network and Service Management},
  year    = {2023},
  note    = {Cyber Security Learning Environment (CSLE) platform; arXiv:2301.06085},
  url     = {https://arxiv.org/abs/2301.06085}
}

@article{landauer2020kyoushi,
  author  = {Landauer, Max and Skopik, Florian and Wurzenberger, Markus and Hotwagner, Wolfgang and Rauber, Andreas},
  title   = {Have It Your Way: Generating Customized Log Datasets with a Model-Driven Simulation Testbed},
  journal = {IEEE Transactions on Reliability},
  year    = {2021},
  volume  = {70},
  number  = {1},
  pages   = {402--415},
  month   = mar,
  doi     = {10.1109/TR.2020.3031317},
  note    = {Kyoushi Testbed Environment / AIT Cyber Range}
}

@article{singer2025incalmo,
  author  = {Singer, Brian and Lucas, Keane and Adiga, Lakshmi and Jain, Meghna and Bauer, Lujo and Sekar, Vyas},
  title   = {{Incalmo}: An Autonomous {LLM}-assisted System for Red Teaming Multi-Host Networks},
  journal = {arXiv preprint arXiv:2501.16466},
  year    = {2025},
  note    = {Introduces the MHBench benchmark; v4, November 2025},
  url     = {https://arxiv.org/abs/2501.16466}
}

@techreport{updyke2018ghosts,
  author      = {Updyke, Dustin and Dobson, Geoffrey and Podnar, Thomas and Osterritter, Luke and Earl, Benjamin and Cerini, Adam},
  title       = {{GHOSTS} in the Machine: A Framework for Cyber-Warfare Exercise {NPC} Simulation},
  institution = {Software Engineering Institute, Carnegie Mellon University},
  year        = {2018},
  number      = {CMU/SEI-2018-TR-005},
  doi         = {10.1184/R1/12363029.v1},
  url         = {https://insights.sei.cmu.edu/library/ghosts-in-the-machine-a-framework-for-cyber-warfare-exercise-npc-simulation/}
}

@inproceedings{rigaki2023meme,
  author    = {Rigaki, Maria and Garcia, Sebastian},
  title     = {The Power of {MEME}: Adversarial Malware Creation with Model-Based Reinforcement Learning},
  booktitle = {Proceedings of the 28th European Symposium on Research in Computer Security (ESORICS)},
  series    = {Lecture Notes in Computer Science},
  year      = {2023},
  doi       = {10.1007/978-3-031-51482-1_3},
  url       = {https://arxiv.org/abs/2308.16562}
}

@inproceedings{fergusonwalter2019tularosa,
  author    = {Ferguson-Walter, Kimberly and Shade, Temmie and Rogers, Andrew and Niedbala, Elizabeth M. and Trumbo, Michael C. and Nauer, Kevin S. and Divis, Kristin M. and Jones, Aaron P. and Combs, Angela and Abbott, Robert G.},
  title     = {The {Tularosa} Study: An Experimental Design and Implementation to Quantify the Effectiveness of Cyber Deception},
  booktitle = {Proceedings of the 52nd Hawaii International Conference on System Sciences (HICSS)},
  year      = {2019},
  pages     = {1--10},
  doi       = {10.24251/HICSS.2019.874}
}

@inproceedings{fergusonwalter2021decoy,
  author    = {Ferguson-Walter, Kimberly J. and Major, Maxine M. and Johnson, Chelsea K. and Muhleman, Daniel H.},
  title     = {Examining the Efficacy of Decoy-based and Psychological Cyber Deception},
  booktitle = {Proceedings of the 30th USENIX Security Symposium (USENIX Security)},
  year      = {2021},
  pages     = {1127--1144}
}

@inproceedings{vetterl2018bitterharvest,
  author    = {Vetterl, Alexander and Clayton, Richard},
  title     = {Bitter Harvest: Systematically Fingerprinting Low- and Medium-interaction Honeypots at Internet Scale},
  booktitle = {Proceedings of the 12th USENIX Workshop on Offensive Technologies (WOOT)},
  year      = {2018}
}

@inproceedings{antonakakis2017mirai,
  author    = {Antonakakis, Manos and April, Tim and Bailey, Michael and Bernhard, Matt and Bursztein, Elie and Cochran, Jaime and Durumeric, Zakir and Halderman, J. Alex and Invernizzi, Luca and Kallitsis, Michalis and Kumar, Deepak and Lever, Chaz and Ma, Zane and Mason, Joshua and Menscher, Damian and Seaman, Chad and Sullivan, Nick and Thomas, Kurt and Zhou, Yi},
  title     = {Understanding the {Mirai} Botnet},
  booktitle = {Proceedings of the 26th USENIX Security Symposium (USENIX Security)},
  year      = {2017},
  pages     = {1093--1110}
}

@inproceedings{barron2017picky,
  author    = {Barron, Timothy and Nikiforakis, Nick},
  title     = {Picky Attackers: Quantifying the Role of System Properties on Intruder Behavior},
  booktitle = {Proceedings of the 33rd Annual Computer Security Applications Conference (ACSAC)},
  year      = {2017},
  pages     = {387--398}
}

@misc{cve20236553,
  author       = {{National Vulnerability Database}},
  title        = {{CVE-2023-6553}: {WordPress} Backup Migration Plugin Remote Code Execution},
  year         = {2023},
  howpublished = {NVD entry; CVSS 9.8; affects Backup Migration plugin versions $\le$ 1.3.7; patched in 1.3.8 (Dec 6, 2023)},
  url          = {https://nvd.nist.gov/vuln/detail/CVE-2023-6553}
}

@inproceedings{du2022adaptive,
  title={Learning to play an adaptive cyber deception game},
  author={Du, Yinuo and Song, Zimeng and Milani, Stephanie and Gonzalez, Cleotilde and Fang, Fei},
  booktitle={Proceedings of the 21st International Conference on Autonomous Agents and Multiagent Systems (AAMAS 2022) Workshop},
  pages={1--6},
  year={2022}
}

@article{prebot2023learning,
  title={Learning about simulated adversaries from human defenders using interactive cyber-defense games},
  author={Prebot, Baptiste and Du, Yinuo and Gonzalez, Cleotilde},
  journal={Journal of Cybersecurity},
  volume={9},
  number={1},
  pages={tyad022},
  year={2023},
  publisher={Oxford University Press}
}

@inproceedings{sayed2023honeypot,
  title={Honeypot allocation for cyber deception in dynamic tactical networks: A game theoretic approach},
  author={Sayed, Md Abu and Anwar, Ahmed H. and Kiekintveld, Christopher and Kamhoua, Charles},
  booktitle={International Conference on Decision and Game Theory for Security (GameSec)},
  pages={195--214},
  year={2023},
  organization={Springer}
}

@article{albrecht2018autonomous,
  author  = {Albrecht, Stefano V. and Stone, Peter},
  title   = {Autonomous Agents Modelling Other Agents: A Comprehensive Survey and Open Problems},
  journal = {Artificial Intelligence},
  volume  = {258},
  pages   = {66--95},
  year    = {2018},
  publisher = {Elsevier},
  doi     = {10.1016/j.artint.2018.01.002},
  url     = {https://arxiv.org/abs/1709.08071}
}

@article{gmytrasiewicz2005framework,
  author  = {Gmytrasiewicz, Piotr J. and Doshi, Prashant},
  title   = {A Framework for Sequential Planning in Multi-Agent Settings},
  journal = {Journal of Artificial Intelligence Research},
  volume  = {24},
  pages   = {49--79},
  year    = {2005},
  url     = {https://jair.org/index.php/jair/article/view/10414}
}

@inproceedings{nguyen2013analyzing,
  author    = {Nguyen, Thanh H. and Yang, Rong and Azaria, Amos and Kraus, Sarit and Tambe, Milind},
  title     = {Analyzing the Effectiveness of Adversary Modeling in Security Games},
  booktitle = {Proceedings of the 27th AAAI Conference on Artificial Intelligence (AAAI)},
  year      = {2013},
  pages     = {718--724},
  doi       = {10.1609/aaai.v27i1.8599}
}

@inproceedings{he2016opponent,
  author    = {He, He and Boyd-Graber, Jordan and Kwok, Kevin and Daum{\'e} III, Hal},
  title     = {Opponent Modeling in Deep Reinforcement Learning},
  booktitle = {Proceedings of the 33rd International Conference on Machine Learning (ICML)},
  year      = {2016},
  pages     = {1804--1813},
  series    = {PMLR},
  volume    = {48},
  url       = {https://arxiv.org/abs/1609.05559}
}

@inproceedings{grover2018learning,
  author    = {Grover, Aditya and Al-Shedivat, Maruan and Gupta, Jayesh K. and Burda, Yuri and Edwards, Harrison},
  title     = {Learning Policy Representations in Multiagent Systems},
  booktitle = {Proceedings of the 35th International Conference on Machine Learning (ICML)},
  year      = {2018},
  pages     = {1802--1811},
  series    = {PMLR},
  volume    = {80},
  url       = {https://arxiv.org/abs/1806.06464}
}

@inproceedings{rabinowitz2018machine,
  author    = {Rabinowitz, Neil C. and Perbet, Frank and Song, H. Francis and Zhang, Chiyuan and Eslami, S. M. Ali and Botvinick, Matthew},
  title     = {Machine Theory of Mind},
  booktitle = {Proceedings of the 35th International Conference on Machine Learning (ICML)},
  year      = {2018},
  pages     = {4218--4227},
  series    = {PMLR},
  volume    = {80},
  url       = {https://arxiv.org/abs/1802.07740}
}

@inproceedings{papoudakis2021agent,
  author    = {Papoudakis, Georgios and Christianos, Filippos and Albrecht, Stefano V.},
  title     = {Agent Modelling under Partial Observability for Deep Reinforcement Learning},
  booktitle = {Advances in Neural Information Processing Systems (NeurIPS)},
  year      = {2021},
  volume    = {34},
  url       = {https://arxiv.org/abs/2006.09447}
}

@inproceedings{jing2024towards,
  author    = {Jing, Yuheng and Li, Kai and Liu, Bingyun and Zang, Yifan and Fu, Haobo and Fu, Qiang and Xing, Junliang and Cheng, Jian},
  title     = {Towards Offline Opponent Modeling with In-Context Learning},
  booktitle = {Proceedings of the 12th International Conference on Learning Representations (ICLR)},
  year      = {2024},
  url       = {https://openreview.net/forum?id=2SwHngthig}
}

@article{li2021cygil,
  author  = {Li, Li and Fayad, Raed and Taylor, Adrian},
  title   = {{CyGIL}: A Cyber Gym for Training Autonomous Agents over Emulated Network Systems},
  journal = {arXiv preprint arXiv:2109.03331},
  year    = {2021},
  url     = {https://arxiv.org/abs/2109.03331}
}

@inproceedings{foley2022autonomous,
  author    = {Foley, Myles and Hicks, Chris and Highnam, Kate and Mavroudis, Vasilios},
  title     = {Autonomous Network Defence using Reinforcement Learning},
  booktitle = {Proceedings of the 2022 ACM Asia Conference on Computer and Communications Security (ASIA CCS)},
  year      = {2022},
  pages     = {1252--1254},
  doi       = {10.1145/3488932.3527286}
}

@article{singh2025hierarchical,
  author  = {Singh, Aditya Vikram and Rathbun, Ethan and Graham, Emma and Oakley, Lisa and Boboila, Simona and Oprea, Alina and Chin, Peter},
  title   = {Hierarchical Multi-agent Reinforcement Learning for Cyber Network Defense},
  journal = {arXiv preprint arXiv:2410.17351},
  year    = {2025},
  url     = {https://arxiv.org/abs/2410.17351}
}

@inproceedings{cobbe2019quantifying,
  author    = {Cobbe, Karl and Klimov, Oleg and Hesse, Chris and Kim, Taehoon and Schulman, John},
  title     = {Quantifying Generalization in Reinforcement Learning},
  booktitle = {Proceedings of the 36th International Conference on Machine Learning (ICML)},
  year      = {2019},
  pages     = {1282--1289},
  series    = {PMLR},
  volume    = {97},
  url       = {https://arxiv.org/abs/1812.02341}
}

@article{nguyen2022deep,
  author  = {Nguyen, Thanh Thi and Reddi, Vijay Janapa},
  title   = {Deep Reinforcement Learning for Cyber Security},
  journal = {IEEE Transactions on Neural Networks and Learning Systems},
  volume  = {34},
  number  = {8},
  pages   = {3779--3795},
  year    = {2023},
  doi     = {10.1109/TNNLS.2021.3121870}
}

\appendices

\section{Notation}

\begin{table}[t]
\centering
\caption{Notation summary.}
\label{tab:notation}
\small
\begin{tabular}{l p{0.62\columnwidth}}
\toprule
\textbf{Symbol} & \textbf{Description} \\
\midrule
$o_t$ & Observation at time $t$ (per-host telemetry vector) \\
$a_t$ & Defender action at time $t$ \\
$r_t$ & Scalar reward at time $t$ \\
$z_t$ & Controllable stochastic latent (action-conditioned) \\
$u_t$ & Exogenous stochastic latent (action-independent) \\
$\tau_t$ & Opponent embedding at time $t$ \\
$h_t$ & Recurrent context aggregator for $z_t$ \\
$f_\theta, g_\eta$ & Encoder, decoder \\
$f_\xi$ & Opponent encoder \\
$h_\zeta$ & Reward predictor \\
$\pi$ & Defender policy (actor) \\
$v$ & Critic / value function \\
$\beta_z, \beta_u$ & KL coefficients in the world-model loss \\
$\alpha, \beta$ & Policy and opponent loss weights \\
$E$ & Episodes collected per training iteration \\
$H$ & Imagination horizon (rollout length) \\
$K$ & Sliding window length for opponent encoder \\
$\gamma$ & Discount factor \\
\bottomrule
\end{tabular}
\end{table}

\section{Hyperparameter Specification}\label{app:hyperparameters}

Table~\ref{tab:hyperparameters} lists the hyperparameters used to produce the reported numbers in Section~\ref{sec:eval}, with search ranges and selected values. Selected values minimize episode loss on a held-out validation set drawn from the same testbed; the bounds shown were used in a small grid search. Reproduction code in the artifact bundle automatically loads the selected values; the bounds are documented so reviewers can verify that selected values do not sit at the search boundary.

\begin{table*}[t]
\centering
\caption{FOE-Dreamer hyperparameters: search range and selected value.}
\label{tab:hyperparameters}
\small
\setlength{\tabcolsep}{3pt}
\begin{tabular}{@{}lcc@{}}
\toprule
\textbf{Hyperparameter} & \shortstack{\textbf{Range}\\\textbf{(lower, upper)}} & \textbf{Selected} \\
\midrule
Controllable latent dim ($z_t$) & 32, 256 & 128 \\
Exogenous latent dim ($u_t$) & 32, 256 & 128 \\
Opponent embedding dim ($\tau$) & 16, 128 & 64 \\
Opponent sliding window ($K$) & 5, 20 & 10 \\
Imagination horizon ($H$) & 10, 50 & 30 \\
KL coefficient $\beta_z$ & 0.1, 2.0 & 1.0 \\
KL coefficient $\beta_u$ & 0.1, 2.0 & 0.5 \\
Policy loss weight $\alpha$ & 0.1, 1.0 & 0.5 \\
Opponent loss weight $\beta$ & 0.1, 1.0 & 0.3 \\
Learning rate (Adam) & $1\!\times\!10^{-5}$, $1\!\times\!10^{-3}$ & $3\!\times\!10^{-4}$ \\
Minibatch size & 32, 256 & 64 \\
Replay buffer capacity & $10^4$, $10^6$ & $10^5$ \\
Discount $\gamma$ & 0.95, 0.999 & 0.99 \\
Episodes per iteration ($E$) & 4, 32 & 8 \\
Episode length (timesteps) & --- & 100 \\
Polling interval $\Delta t$ (seconds) & --- & 5 \\
Random seeds per configuration & --- & 5 \\
Training budget (wall-clock days) & --- & 3 \\
\bottomrule
\end{tabular}
\end{table*}

The fixed values (episode length, polling interval, seeds, training budget) are determined by the testbed and resource budget rather than by tuning. Hyperparameters tuned via grid search were explored over a small grid; selected values were chosen from configurations within one standard deviation of the best validation loss. The actor, critic, world-model encoder/decoder, and opponent encoder use small MLP and recurrent architectures; the optimizer is Adam with gradient clipping. The recurrent world model and imagination-based actor--critic follow the Dreamer framework \cite{hafner2021dreamerv2}; Table~\ref{tab:hyperparameters} specifies the FOE-Dreamer dimensions and training settings.

\newpage
\section*{LLM Usage Statement}

LLMs were used for editorial purposes in this manuscript, and all outputs were inspected by the authors to ensure accuracy and originality. Specifically, LLM assistance was limited to prose polishing, sentence-level rephrasing for clarity, and consistency checking. All intellectual content, technical claims, system design, experimental methodology, results, and analysis are the authors' own work. No code, equations, proofs, or empirical results were generated by LLMs.

\end{document}